\documentclass[]{iscram}

\usepackage[utf8]{inputenc}

\usepackage{lipsum}
\usepackage{tikz}
\usepackage{fancyvrb}
\usepackage{listings}
\usepackage{enumitem}
\usepackage{tcolorbox}
\usepackage{multicol}
\usepackage{multirow} 
\usepackage{makecell} 
\usepackage{siunitx}
\usepackage{tikzpagenodes}
\usepackage{xcolor}
\usepackage{makecell}
\usepackage{hyperref}

\lstdefinestyle{common}{
  xleftmargin=.5em,
  xrightmargin=.5em,
  frame=single,framesep=.5em,framerule=0pt,
  fancyvrb=true,
  basicstyle=\ttfamily,
  keywordstyle=\color{cyan!50!blue!75!black}\bfseries,
  commentstyle=\color{red!50!black}\itshape,
  stringstyle=\ttfamily\color{green!50!black},
  numbers=none,
  showspaces=false,
  showstringspaces=false,
  fontadjust=true,
  keepspaces=true,
  flexiblecolumns=true,
  emphstyle=\color{red},
}
\lstdefinestyle{TeX}{
  style=common,
  backgroundcolor=\color{blue!5},
  aboveskip=5pt,
  belowskip=5pt,
  language=[LaTeX]TeX,
  moretexcs={
    abstract, addbibresource, iscramset, keywords, mainmatter,
    maketitle, printbibliography, subsection, subsubsection, url,
    urldef, href, includegraphics, ldots, parencite, citeauthor,
    citeyear, citetitle, midrule, toprule, bottomrule
  },
  fancyvrb=true,
}

\lstdefinestyle{console}{
  style=common,
  backgroundcolor=\color{gray!10},
  aboveskip=5pt,
  belowskip=5pt,
}

\newlist{options}{description}{1}
\setlist[options]{%
  beginpenalty=10000,%
  itemsep=.5\parskip plus .3\parskip minus .2\parskip,
  parsep=.5\parskip plus .3\parskip minus .2\parskip,
  topsep=.5\parskip plus .3\parskip minus .2\parskip,
  partopsep=.5\parskip plus .3\parskip minus .2\parskip,
  style=nextline,labelindent=1em,%
  font=\normalfont\ttfamily}

\colorlet{macro color}{cyan!50!blue!75!black}
\colorlet{option color}{red!50!black}
\colorlet{generic color}{green!40!black}

\newtcolorbox{pseudoTeX}{colback=blue!5,colframe=blue!5,before=\nobreak}
\let\LaTeXorig\LaTeX
\renewcommand\LaTeX{\bgroup\fontfamily{lmr}\selectfont\upshape\LaTeXorig\egroup}

\urldef{\jacobemail}\url{jativo@horizon.csueastbay.edu}
\urldef{\bharaniemail}\url{bharanibala@ksu.edu}
\urldef{\anhemail}\url{anhtranst@gmail.com}
\urldef{\khushbooemail}\url{kgupta27@uic.edu}
\urldef{\hongminemail}\url{hongmin.li@csueastbay.edu}
\urldef{\dcarageaemail}\url{dcaragea@ksu.edu}
\urldef{\ccarageaemail}\url{cornelia@uic.edu}

\newcommand{\CSUEBaffil}{
  Department of Computer Science \\
  California State University, East Bay \\
  Hayward CA 94542, USA
}

\newcommand{\KSUaffil}{
  Department of Computer Science \\
  Kansas State University
}

\newcommand{\UICaffil}{
  Department of Computer Science \\
  University of Illinois at Chicago
}

\newcommand{\equalcontrib}{\thanks{These authors contributed equally to this work.}}

\usetikzlibrary{fit}

\iscramset{
  WiPe Paper 2026={Social Media \& Crisis Communication: narratives, signals, and sentiments},
  title={LLM-guided Semi-Supervised Approaches for Social Media Crisis Data Classification},
  short title={LLM-guided Semi-Supervised for Crisis},
  author={
    short name={Ativo et al.},
    full name=Jacob Ativo\equalcontrib,
    affiliation={
      \CSUEBaffil \\
\href{mailto:jativo@horizon.csueastbay.edu}{\jacobemail}
    },
},
author={
    short name={Balasubramaniyam},
    full name=Bharaneeshwar Balasubramaniyam\footnotemark[1],
    affiliation={
      \KSUaffil \\
      \href{mailto:bharanibala@ksu.edu}{\bharaniemail}
    },
},
author={
    short name={Tran},
    full name=Anh Tran\footnotemark[1],
    affiliation={
      Independent Researcher \\
      \href{mailto:anhtranst@gmail.com}{\anhemail}
    },
},
author={
    short name={Gupta},
    full name=Khushboo Gupta,
    affiliation={
      \UICaffil \\
      \href{mailto:kgupta27@uic.edu}{\khushbooemail}
    },
},
author={
    short name={Li},
    full name=Hongmin Li,
    affiliation={
      \CSUEBaffil \\
      \href{mailto:hongmin.li@csueastbay.edu}{\hongminemail}
    },
},
author={
    short name={D. Caragea},
    full name=Doina Caragea,
    affiliation={
      \KSUaffil \\
      \href{mailto:dcaragea@ksu.edu}{\dcarageaemail}
    },
},
author={
    short name={C. Caragea},
    full name=Cornelia Caragea,
    affiliation={
      \UICaffil \\
      \href{mailto:cornelia@uic.edu}{\ccarageaemail}
    },
},
}

\begin{document}

\maketitle

\abstract{
Semi-supervised learning approaches have been investigated as a means to enhance the analysis of social media data in disaster management contexts. In this work, we present the first empirical evaluation of large language model (LLM) guided semi-supervised learning for crisis related tweet classification. We compare two recent LLM assisted semi-supervised methods, VerifyMatch and LLM guided Co-Training ({ \sc LG-CoTrain}), against established semi-supervised baselines. Our results show that { \sc LG-CoTrain} significantly outperforms classical semi-supervised approaches in low resource settings with 5, 10 and 25 labeled examples per class, achieving the highest averaged Macro F1 across events. VerifyMatch achieves competitive performance while also demonstrating strong calibration properties. As the number of labeled examples increases, the performance gap narrows and Self Training emerges as a strong baseline. We further observe that compact semi-supervised models can, in some cases, outperform very large LLMs operating in zero-shot settings. This finding highlights the potential of transferring knowledge from LLMs into smaller and more deployable models through LLM guided semi-supervised learning, offering a practical pathway for real world disaster response applications. Our project repository on Github is \href{https://github.com/deeplearning-lab-csueb/LLM-guided-SSL-Crisis-Tweets-Classification/tree/main}{here}. 
}


\section{Introduction}

During emergency events, individuals increasingly turn to social media platforms such as X (formerly Twitter), Reddit, and Instagram to seek information and share updates. From a communication perspective, these platforms function bidirectionally: authorities disseminate critical disaster response information (e.g., warnings or evacuation orders) to the public, while the public also provides firsthand reports and situational updates that can be mined to enhance situational awareness \parencite{reuter2018_survey, DBLP:phd/dnb/Reuter22a, jeroen2021}. Consequently, both researchers and practitioners recognize the substantial value of such user-generated content for crisis response. However, effectively integrating social media streams into real-time operations remains challenging due to information overload characterized by high volume, velocity, and varying levels of veracity \parencite{purohit2025engagemobilizeunderstandingevolving}.

To address these challenges, extensive research over the past decade has focused on applying Machine Learning (ML) and Natural Language Processing (NLP) techniques to automatically classify social media data into actionable categories, such as infrastructure damage or requests for rescue. A wide range of ML models have been proposed for these classification tasks, including statistical learning approaches and supervised deep learning models \parencite{starbirdPHV2010,ImranECDM13,CarageaSSNT14,NguyenAJSIM16, BurelAlani2018, DBLP:conf/iscram/KerstenKWK19,DBLP:journals/ipm/GhafarianY20}. However, supervised models typically require substantial amounts of high-quality human-labeled data to achieve strong performance, which is often scarce in the time-sensitive context of disaster response.

To mitigate this limitation, researchers have explored domain adaptation, transfer learning, and semi-supervised learning approaches. Domain adaptation methods leverage labeled data from previous disaster events to alleviate label scarcity in newly emerging events \parencite{LiJCCM2017,imran2028domainadaptation}. In contrast, semi-supervised learning methods aim to train effective models by combining a small amount of labeled data with a large volume of unlabeled data through pseudo-labeling strategies. In a typical teacher–student semi-supervised learning framework, a teacher model trained on the limited labeled data first generates pseudo-labels for the unlabeled instances, and these pseudo-labeled examples are subsequently used to train a student model \parencite{Li-iscram-21, zou2023crisismatch, gupta_2025_calibrated}. In general, the performance of semi-supervised approaches depends heavily on the quality of the pseudo-labels. Therefore, a central research question in semi-supervised learning is how to effectively leverage unlabeled data to generate high-quality pseudo-labels that improve downstream model performance.

With the rapid advancement of Large Language Models (LLMs), recent studies have explored leveraging LLMs to improve pseudo-labeling of semi-supervised models built on smaller pre-trained language models such as BERT \parencite{DBLP:journals/corr/abs-1810-04805-bert}, particularly for text classification tasks \parencite{park_2024_verifymatch, rahman_caragea_2025_llm}. In the context of social media crisis data analysis, there has been a surge of work employing LLMs in 
zero-shot (i.e., making predictions using the LLM without task-specific labeled examples), few-shot (i.e., provide the LLM a small number of labeled examples in the prompt), and fine-tuning (i.e., updating a small sized LLM model parameters on task-specific labeled data)
to identify informative social media content for disaster management \parencite{imran2024-openai, Soudabeh-Caragea-2024,mcdaniel2024-zeroshot-crisisbench,yin2025-crisisSense,shrestha_crisis_tweets_2025-thesis,DBLP:conf/cogsima/SalfingerS24, lei2025harnessing, guo_2025_asonam}. However, to the best of our knowledge, no prior work has investigated semi-supervised models guided by LLMs in the crisis domain.

To this end, we study two BERT-based semi-supervised models enhanced with LLM-generated pseudo-labels for social media crisis classification: (1) VerifyMatch \parencite{park_2024_verifymatch}, originally proposed for Natural Language Inference (NLI), and (2) LLM-guided Co-training ({\sc LG-CoTrain}) \parencite{rahman_caragea_2025_llm}, designed for general text classification. 
Following the experimental protocol of \textcite{gupta_2025_calibrated}, which evaluates several semi-supervised methods on 10 disaster events from the HumAID dataset \parencite{alam_2021_humaid}, we experiment with VerifyMatch and {\sc LG-CoTrain}, enhanced with GPT-4o pseudo labels, on the same benchmark.

Specifically, we evaluate the performance of the VerifyMatch and {\sc LG-CoTrain} approaches using the Macro-F1 score, as well as the Expected Calibration Error (ECE), and compare the results with those of the existing baselines from \textcite{gupta_2025_calibrated} to form a more comprehensive study for semi-supervised learning algorithms on social media crisis data classification. 
To summarize, our main contributions are as follows:
\begin{itemize}

    \item We evaluate two semi-supervised approaches, VerifyMatch and {\sc LG-CoTrain}, using zero-shot pseudo-labels generated by GPT-4o on 10 disaster events from the HumAID dataset, a benchmark of disaster-related tweets annotated with humanitarian categories such as damage, injured people, and requests or urgent needs.
    We further compare these models against all semi-supervised methods examined by \textcite{gupta_2025_calibrated}.
    \item Our experimental results show that {\sc LG-CoTrain} significantly outperforms other approaches in low-resource settings (e.g., 5 or 10 labeled examples per category). Moreover, it demonstrates good model calibration. 
    However, as the amount of labeled data increases (e.g., 50 labeled examples per category), the performance gap between {\sc LG-CoTrain} and other semi-supervised models narrows, and Self-training emerges as a competitive baseline that is difficult to surpass.
    \item Semi-supervised models based on smaller pre-trained language models outperform zero-shot GPT-4o only on a subset of disaster events. This may be due to limited and potentially unrepresentative unlabeled data in the HumAID benchmark, including missing class examples in the sampled unlabeled sets—an issue that can also arise in real-world scenarios. Larger and more representative unlabeled datasets could help mitigate these limitations.
\end{itemize}

 Still all these findings highlight the potential of transferring knowledge from LLMs into smaller and more deployable models through LLM guided semi-supervised learning, offering a practical pathway for real world disaster response applications.

\section{Related Work}

There is a vast body of literature on semi-supervised learning (SSL) in machine learning. In this section, we first provide an overview of SSL approaches, and then review prior work applying SSL and large language models (LLMs) to social media disaster data analysis.

\textbf{SSL Overview}.
A wide range of SSL approaches have been proposed for text classification, beginning with the original idea of Self-Training and pseudo-labeling \parencite{scudder1965probability}. In self-training, a teacher model is first trained on limited labeled data and then used to generate pseudo-labels for unlabeled instances. These pseudo-labeled examples are subsequently incorporated into training a student model, often in an iterative manner. Two critical design choices in this framework are (1) how to select pseudo-labeled examples for inclusion in training and (2) whether to use hard labels (the most likely class per example) or soft labels (predicted class probabilities). Incorporating low-confidence pseudo-labels may lead to error propagation and degrade student model performance.

Various pseudo-label selection strategies have been proposed. For example, FixMatch \parencite{sohn2020fixmatch} and related self-training methods adopt fixed confidence thresholds to filter pseudo-labels, while Uncertainty-Aware Self-Training (UST) \parencite{mukherjee2020uncertainty} employs more sophisticated uncertainty estimation techniques grounded in probability theory. However, threshold-based filtering may restrict the student model’s access to potentially informative unlabeled data. To alleviate this limitation, methods such as MixMatch \parencite{berthelot2019mixmatch} and SoftMatch \parencite{chen2023softmatchaddressingquantityqualitytradeoff} were introduced. SoftMatch retains all pseudo-labeled samples but assigns lower weights to low-confidence instances during training, thereby balancing data quantity and label quality. MixMatch similarly leverages soft pseudo-labels and further incorporates MixUp \parencite{zhang2017mixup}, which interpolates pseudo-labeled and human-annotated examples to generate smoother and potentially higher-quality training signals. 

AUM-based Self-Training (AUM-ST) \parencite{sosea2022leveraging} takes a different perspective by filtering low-quality pseudo-labeled examples through tracking training dynamics using the Area Under the Margin (AUM). Building upon this idea, AUM-based MixUp in Self-Training (AUM-ST-Mixup) \parencite{gupta_2025_calibrated} integrates MixUp and additional confidence-tracking mechanisms on top of AUM-ST to further enhance pseudo-label reliability.
Confidence-based Mixup in Self-Training (Conf-ST-Mixup) \parencite{gupta_2025_calibrated} enhances pseudo-labeling by defining prediction confidence as the probability gap between the top two classes, where a larger gap indicates higher confidence, allowing the model to distinguish easy-to-learn (reliable) from hard-to-learn (ambiguous) samples. It applies mixup across labeled, high-confidence, and low-confidence pseudo-labeled data to regularize training, reduce error propagation, and promote smoother decision boundaries.

Despite these improvements, methods relying on a single model for pseudo-label generation remain vulnerable to reinforcing incorrect high-confidence predictions, particularly in early training stages \parencite{rahman_caragea_2025_llm}. To mitigate this issue, VerifyMatch \parencite{park_2024_verifymatch} incorporates LLM-generated pseudo-labels alongside a verifier model, enabling cross-validation of pseudo-label quality. Combined with MixUp, VerifyMatch achieves competitive performance in low-resource settings. Finally, LLM-guided Co-Training ({\sc LG-CoTrain}) \parencite{rahman_caragea_2025_llm} integrates LLM-generated pseudo-labels within a dual-model co-training framework, where two models iteratively learn from each other while incorporating LLM guidance. Unlike MixUp-based approaches, {\sc LG-CoTrain} retains all pseudo-labeled data without modification. {\sc LG-CoTrain} outperforms the zero-shot Phi-3 and other SSL approaches and achieves state-of-the-art performance on four out of five text classification benchmark datasets.

\textbf{SSL for Social Media Crisis Data Analysis}. 
Several studies have applied SSL techniques to social media crisis data analysis. For example, \textcite{alam2018graph} proposed a graph-based semi-supervised CNN model for Twitter data from two disaster events. 
\textcite{Li-iscram-21} applied self-training with BERT and CNN models to the CrisisLexT6 and CrisisLexT26 datasets \parencite{OlteanuCDV14, Olteanu2015}, which contain tweets from various disaster events. CrisisLexT6 is annotated for whether tweets are related to the disaster, while CrisisLexT26 includes coarse- and fine-grained humanitarian labels similar to those in the HumAID dataset.
\textcite{sirbu2022multimodal} extended FixMatch by incorporating soft-labeling for multimodal disaster tweet classification (text and images) on the CrisisMMD dataset \parencite{crisismmd_2018_icwsm}. \textcite{zou2023crisismatch} proposes CrisisMatch which differs from MixMatch by using hard pseudo-labeling for entropy maximization instead of sharpening for text classification on the HumAID dataset. Meanwhile, \textcite{zou2023decrisismb} proposes a novel approach by using memory bank --- DeCrisisMB \parencite{zou2023decrisismb} --- to addresses the bias in SSL which assigns disproportionate pseudo-labels for more occurring instances in highly imbalanced datasets such as crisis-related tweet classification.

More recently, \textcite{gupta_2025_calibrated} proposed Confidence-based and AUM-based MixUp with Self-Training (AUM-ST-MixUp) and conducted a systematic evaluation across 10 disaster events in the HumAID dataset, comparing these methods with several SSL baselines discussed above. In addition, \textcite{gupta_2025_calibrated} employed Expected Calibration Error (ECE) to measure model calibration. ECE quantifies the extent to which predicted probabilities align with observed outcomes. A well-calibrated model produces confidence estimates that align with empirical correctness rates; for example, predictions assigned 85\% confidence should be correct approximately 85\% of the time. Calibration is particularly important for interpreting model outputs and supporting reliable decision-making in high-stakes contexts such as disaster response.

\textbf{LLMs for Social Media Crisis Data Analysis}.
With the rapid development of LLMs, increasing attention has been devoted to applying them to social media data analysis \parencite{lei2025harnessing, sanchez_2025_llmcifsc}. For instance, 
\textcite{guo_2025_asonam} compared fine-tuned Llama 3.2 11B models with zero-shot GPT-4o and prior approaches, demonstrating that fine-tuned Llama models achieve state-of-the-art performance on multimodal crisis classification using the CrisisMMD dataset. Most closely related to our work is \textcite{imran2024-openai}, which systematically evaluated the robustness of several LLMs—including GPT-3.5, GPT-4, GPT-4o, Llama-2 13B, Llama-3 8B, and Mistral 7B—on the HumAID dataset under zero-shot and few-shot settings. Their results indicate that few-shot prompting does not consistently improve performance, and GPT-4 achieved the strongest overall results among the evaluated models. 

In this work, we compare LLM-guided SSL approaches against the SSL methods evaluated in \textcite{gupta_2025_calibrated} with respect to model prediction accuracy, as well as calibration error (ECE metric).

\section{Dataset}
We use the same 10 disaster events from the HumAID dataset as our benchmark following \textcite{gupta_2025_calibrated}, as shown in Table~\ref{tab:dataset_stats}. HumAID is a human-annotated Twitter dataset comprising 77,196 tweets from 19 disaster events, categorized into 11 classes. Excluding the ``Don't know or can't judge'' category, the remaining 10 primary humanitarian categories are: \textit{1. Caution and advice; 2. Sympathy and support; 3. Requests or urgent needs; 4. Displaced people and evacuations; 5. Injured or dead people; 6. Missing or found people; 7. Infrastructure and utility damage; 8. Rescue, volunteering, or donation effort; 9. Other relevant; 10. Not humanitarian}.

Table~\ref{tab:dataset_stats} and ~\ref{tab:disaster_class_distribution} also reports the statistics of the training, validation, and test splits, number of classes for each event as well as class distribution. For detailed class distributions within each event, we refer readers to \textcite{gupta_2025_calibrated}. Consistent with their experimental setup, we adopt the same data split configuration, in which the training set is further divided into labeled and unlabeled subsets, as described in the Experimental Setup section.

\begin{table*}[t]
\centering

\scriptsize 
\renewcommand{\arraystretch}{1.0} 
\resizebox{\textwidth}{!}{ 

\begin{tabular}{l|r|ccc|cc|cc|cc|cc}
\toprule
Disaster Event/Data Split & C & Train & Val & Test 
    & \multicolumn{2}{c|}{5 lb/cl}
    & \multicolumn{2}{c|}{10 lb/cl} 
    & \multicolumn{2}{c|}{25 lb/cl}
    & \multicolumn{2}{c}{50 lb/cl} \\

\midrule
 &  &  &  &  & $L$ & $U$ & $L$ & $U$ & $L$ & $U$ & $L$ & $U$ \\
\midrule

California Wildfires 2018  & 10 & 5163 & 752 & 1461 & 50 & 5113 & 100 & 5063 & 250 & 4913 & 500 & 4663 \\
Canada Wildfires 2016      &  8 & 1569 & 228 & 445  & 40 & 1529 & 80 & 1489 & 189 & 1380 & 364 & 1205 \\
Cyclone Idai 2019          & 10 & 2753 & 401 & 779  & 50 & 2703 & 100 & 2653 & 238 & 2515 & 453 & 2300 \\
Hurricane Dorian 2019      &  9 & 5329 & 776 & 1508 & 45 & 5284 & 90 & 5239 & 225 & 5104 & 442 & 4887 \\
Hurricane Florence 2018    &  9 & 4384 & 639 & 1241 & 45 & 4339 & 90 & 4294 & 225 & 4159 & 438 & 3946 \\
Hurricane Harvey 2017      &  9 & 6378 & 929 & 1805 & 45 & 6333 & 90 & 6288 & 225 & 6153 & 450 & 5928 \\
Hurricane Irma 2017        &  9 & 6579 & 954 & 1862 & 45 & 6534 & 90 & 6489 & 225 & 6354 & 450 & 6129 \\
Hurricane Maria 2017       &  9 & 5094 & 742 & 1442 & 45 & 5049 & 90 & 5004 & 225 & 4869 & 450 & 4644 \\
Kaikoura Earthquake 2016   &  9 & 1536 & 224 & 435  & 45 & 1491 & 90 & 1446 & 217 & 1319 & 417 & 1119 \\
Kerala Floods 2018         &  9 & 5588 & 814 & 1582 & 45 & 5543 & 90 & 5498 & 225 & 5363 & 439 & 5149 \\

\bottomrule
\end{tabular}
}
\caption{Data distribution and splits under different labels-per-class (lb/cl) settings for each of the 10 events in the HumAID dataset, where $C$ stands for the number of classes, $L$ stands for the number of labeled instances and $U$ stands for the number of unlabeled instances.}
\label{tab:dataset_stats}
\end{table*}

\begin{table*}[ht]
\centering
\small
\begin{tabular}{l|c|cccccccccc}
\midrule
\multirow{2}{*}{\textbf{Disaster Event}} & \multirow{2}{*}{\textbf{Classes}}  & \multicolumn{10}{c}{\textbf{Class Distribution}} \\
\cline{3-12}
\rule{0pt}{3ex} &  & 1 & 2 & 3 & 4 & 5 & 6 & 7 & 8 & 9 & 10 \\
\midrule
California Wildfires 2018 & 10 &  97 & 330 & 55 & 258 & 1362 & 125 & 295 & 991 & 727 & 923 \\
Canada Wildfires 2016 & 8 &  74 & 113 & 14 & 266 & 0 & 0 & 176 & 653 & 218 & 55 \\
Cyclone Idai 2019 & 10 & 62 & 338 & 100 & 40 & 303 & 13 & 248 & 1308 & 285 & 56 \\
Hurricane Dorian 2019 & 9 & 958 & 758 & 125 & 561 & 42 & 0 & 571 & 691 & 1011 & 612 \\
Hurricane Florence 2018 & 9 &  917 & 330 & 38 & 446 & 208 & 0 & 224 & 1034 & 445 & 742 \\
Hurricane Harvey 2017 & 9 &  379 & 444 & 233 & 482 & 488 & 0 & 852 & 1976 & 1237 & 287 \\
Hurricane Irma 2017 & 9 &  429 & 397 & 88 & 528 & 626 & 0 & 1317 & 1113 & 1651 & 430 \\
Hurricane Maria 2017 & 9 &  154 & 470 & 498 & 92 & 211 & 0 & 999 & 1384 & 1097 & 189 \\
Kaikoura Earthquake 2016 & 9 & 345 & 302 & 17 & 61 & 73 & 0 & 218 & 145 & 218 & 157 \\
Kerala Floods 2018 & 9 & 97 & 585 & 413 & 39 & 254 & 0 & 207 & 3005 & 669 & 319 \\
\hline
\end{tabular}
\caption{Disaster events and the corresponding number of classes, number of tweets in train split per event with the class distribution}
\label{tab:disaster_class_distribution}
\end{table*}

\section{Methods}

We study semi-supervised learning (SSL) for crisis-related tweet classification under limited labeled data. 
All methods operate on a shared experimental setting consisting of a small labeled subset 
$\mathcal{D}_L = \{(x_i, y_i)\}_{i=1}^{n_L}$ 
and a larger unlabeled subset 
$\mathcal{D}_U = \{x_j\}_{j=1}^{n_U}$ 
drawn from the same disaster event. 
Our goal is to learn a classifier $f_\theta(x)$ that generalizes well to held-out test data while maintaining calibrated confidence estimates.
Unless otherwise specified, all neural models use BERTweet as the encoder backbone, due to its overall good supervised performance \parencite{gupta_2025_calibrated}, followed by a task-specific classification head.

\vspace{-0.2cm}
\subsection{Supervised Baselines}

We include two supervised baselines to contextualize SSL performance.
\vspace{-0.3cm}
\begin{itemize}
\item \textbf{Limited-Label Supervision - Lower Bound.} 
A BERTweet classifier trained solely on $\mathcal{D}_L$ serves as a lower-bound reference, representing performance achievable without unlabeled data.
\vspace{-0.2cm}
\item \textbf{Full-Supervision - Upper Bound.}
A BERTweet model trained on the complete labeled training split (including $\mathcal{D}_L$ and $\mathcal{D}_U$ subsets) provides an approximate upper bound for in-domain performance (Table~\ref{tab:model_comparison} BERTweet All). 
\end{itemize}
\vspace{-0.3cm}

\subsection{Zero-Shot LLM Baseline}

\textbf{Zero-shot setting}: 
To contextualize LLM-guided SSL performance, we evaluate GPT-4o in a zero-shot classification setting, where the LLM directly predicts class labels for test instances without any labeled training example and no task-specific fine-tuning. This baseline measures the standalone capability of LLMs relative to compact supervised and semi-supervised models. Specifically, we experimented with GPT-4.1, GPT-4o, GPT-4o mini, and GPT-5.1, and among them, GPT-4o consistently achieved better performance on both training and test splits. Therefore, we use GPT-4o to generate pseudo-labels for the entire unlabeled training set. We attempted to reproduce the best overall result obtained with zero-shot GPT-4 by \textcite{imran2024-openai}, however, due to API deprecation, exact replication was not possible. Nevertheless, our zero-shot GPT-4o results are comparable (slightly better) on average to the zero-shot GPT-4o reported by \textcite{imran2024-openai} as shown in Table~\ref{tab:model_comparison}. While GPT-4o-mini produced competitive results, it generated a number of out-of-source (OOS) predictions. 

Prompt engineering: We evaluate three prompts with GPT-4o on the validation splits of all 10 events, ranging from simple category definitions from the HumAID dataset to more detailed descriptions. As performance remains nearly identical across prompts (Macro-F1 range: 0.601--0.613) with the simplest version being the best, we adopt the simplest version. The detailed prompt

\vspace{-0.1cm}
\subsection{Classical Semi-Supervised Learning}

We also used representative SSL approaches previously evaluated on the HumAID dataset \parencite{gupta_2025_calibrated}.
\vspace{-0.3cm}
\begin{itemize}
\item \textbf{Self-Training (ST).}
A teacher model trained on $\mathcal{D}_L$ generates pseudo-labels $\hat{y}_j$ for unlabeled examples in $\mathcal{D}_U$. 
High-confidence pseudo-labeled instances are iteratively incorporated into training.
\vspace{-0.2cm}
\item \textbf{Uncertainty-Aware Self-Training (UST).}
UST refines pseudo-label selection by incorporating uncertainty estimation, reducing the influence of noisy high-confidence predictions.
\vspace{-0.2cm}
\item \textbf{MixMatch.}
MixMatch integrates soft pseudo-labeling with MixUp interpolation between labeled and unlabeled examples, encouraging smoother decision boundaries.
\vspace{-0.2cm}
\item \textbf{AUM-based Self-Training (AUM-ST) and AUM-ST-MixUp.}
AUM-ST tracks training dynamics via Area Under the Margin (AUM) to filter unreliable pseudo-labels and AUM-ST-MixUp combines this filtering with MixUp-based regularization. 
\vspace{-0.2cm}
\item \textbf{Confidence-based MixUp in Self-Training (Conf-ST-MixUp).} Conf-ST-MixUp enhances pseudo-labeling by defining prediction confidence as the probability gap between the top two classes, where a larger gap indicates a more reliable pseudo-label. It then applies MixUp across labeled, high-confidence, and low-confidence pseudo-labeled data to regularize training and reduce error propagation.
\end{itemize}
\vspace{-0.3cm}

\subsection{LLM-Guided Semi-Supervised Learning}

We investigate two SSL frameworks that incorporate zero-shot pseudo-labels generated by an LLM, in our case, GPT-4o.
These pseudo-labels are used to augment or guide the training of smaller, task-specific models.
\vspace{-0.3cm}
\begin{itemize}
\item \textbf{VerifyMatch.}
VerifyMatch integrates LLM-generated pseudo-labels with a verifier model that cross-validates predictions before incorporating them into training. Combined with confidence-aware MixUp, this mechanism aims to reduce confirmation bias and mitigate overconfident errors, two common issues in pseudo-labeling. Confirmation bias refers to the tendency of a model to reinforce its own incorrect pseudo-labels during self-training, for example, repeatedly assigning the same wrong label to similar inputs. Overconfident errors refer to incorrect predictions made with high confidence, for example, assigning a wrong label with near-certain probability.
\vspace{-0.2cm}
\item \textbf{LLM-Guided Co-Training ({\sc LG-CoTrain}).}
{\sc LG-CoTrain} employs a dual-model co-training architecture in which two classifiers iteratively exchange pseudo-labels while incorporating LLM guidance.
Unlike MixUp-based methods, {\sc LG-CoTrain} retains all pseudo-labeled examples and relies on cross-model agreement to stabilize learning.
This framework is particularly effective in extremely low-resource settings, where model-generated pseudo-labels alone may be unreliable.
\end{itemize}

Overall, the evaluated methods differ along three dimensions:
(1) pseudo-label generation source (model-based vs LLM-based),
(2) pseudo-label filtering strategy (confidence thresholding, uncertainty-aware weighting, verification, or co-training),
and (3) representation regularization (none, MixUp, or cross-model consistency).
This structured comparison enables analysis of both predictive performance and calibration behavior under label scarcity.

\section{Experimental Setup}
We run experiments with all the approaches described in the Methods section. 
For each event, we use the same train/validation/test splits as \textcite{gupta_2025_calibrated}. We simulate low-resource settings by selecting a fixed number of labeled examples per class (lb/cl) from the training split. We evaluate four label budgets: 5, 10, 25, and 50 lb/cl. The remaining training instances are treated as unlabeled and are used by SSL methods according to their respective learning objectives. This split configuration is kept consistent across all methods to ensure a fair comparison.

We use the same metrics as in \textcite{gupta_2025_calibrated}, specifically, Macro-F1 and the Expected Calibration Error (ECE) averaged across the 10 disaster events, to evaluate all methods. Macro-F1 captures balanced performance across classes, while ECE quantifies how well predicted probabilities align with empirical correctness.

Hyperparameter tuning was performed using Weights \& Biases and the Optuna package over learning rate, batch size, number of epochs, and additional some main model-specific parameters. Detailed configurations will be released in the project repository. For Weights \& Biases, we employed Bayesian sweeps to automate the search; however, we observed occasional optimization instability. Prior work \parencite{liu-wang-2021-empirical}
shows that, in low-resource transformer fine-tuning, automated hyperparameter optimization may fail to outperform simple grid-search under limited search budgets due to overfitting and instability. Moreover, repeated tuning on a small validation set can lead to meta-overfitting, where configurations that perform well on the development set do not generalize to test performance or calibration. This may also explain small discrepancies between our results and those reported in \textcite{gupta_2025_calibrated}.

\section{Results and Discussion}
We show our experimental results in Table~\ref{tab:model_comparison} and Table~\ref{tab:ssl-baselines-vs-ours}, as well as Figure~\ref{fig:macro_f1_heatmap}. 
Table~\ref{tab:model_comparison} reports the zero-shot performance of GPT-4o, along with the performance of the fully supervised BERTweet upper bound. Table~\ref{tab:ssl-baselines-vs-ours} summarizes the Macro-F1 and ECE scores averaged across the 10 disaster events under varying label budgets.

\begin{table}[ht]
\centering
\begin{tabular}{ c| c | c |c| c}
\toprule
Model & 
\makecell{Zero-shot GPT-4o \\ \parencite{imran2024-openai}} &
\makecell{Zero-shot GPT-4o \\ train} & 
\makecell{Zero-shot GPT-4o \\ test} & 
\makecell{BERTweet - All \\} \\
\hline
F1 $\uparrow$ & 0.612 & 0.628 & 0.641 & 0.678 \\
ECE $\downarrow$& - & - & - & 0.110 \\
\bottomrule
\end{tabular}
\caption{Performance results for Zero-shot GPT-4o and supervised BERTweet trained on the whole training split}
\label{tab:model_comparison}
\end{table}

\begin{table}[htbp]
\centering
\small
\begin{tabular}{l|cccc|cccc}

\toprule
Method/Metric & \multicolumn{4}{c|}{F1 $\uparrow$ } & \multicolumn{4}{c}{ECE $\downarrow$}   \\

\# Label & 5 & 10 & 25 & 50 & 5 & 10 & 25 & 50  \\

\midrule

BERTweet          & 0.423 & 0.517 & 0.563 & 0.606 & 0.206 & 0.222 & 0.247 & 0.256 \\
\midrule
ST                & 0.448 & 0.548 & 0.625 & \textbf{0.655} & 0.305 & 0.231 & 0.184 & 0.165 \\
UST               & 0.465 & 0.546 & 0.609 & 0.641 & 0.342 & 0.271 & 0.225 & 0.191 \\
MixMatch          & 0.459 & 0.553 & 0.624 & 0.647 & 0.374 & 0.297 & 0.246 & 0.228 \\
AUM-ST            & 0.424 & 0.505 & 0.572 & 0.595 & 0.264 & 0.206 & 0.204 & 0.194 \\
Conf-ST-MixUp     & 0.421 & 0.533 & 0.623 & 0.643 & 0.408 & 0.321 & 0.244 & 0.246 \\
AUM-ST-MixUp      & 0.476 & 0.532 & 0.611 & 0.639 & 0.190 & \textbf{0.069} & \textbf{0.057} & \textbf{0.064} \\

\midrule

VerifyMatch       & 0.463 & 0.549 & 0.616 & 0.644 & \textbf{0.127} & 0.086 & 0.083 & 0.100 \\
{\sc LG-CoTrain}  & \textbf{0.608} & \textbf{0.619} & \textbf{0.631} & 0.645 & 0.174 & 0.160 & 0.122 & 0.108 \\

\bottomrule
\end{tabular}
\caption{BERTweet and SSL performance on the 10 HumAID disaster events. The results are reported in terms of Macro-F1 and ECE values averaged over the 10 events in the dataset. The best result for each setup is highlighted in bold.}
\label{tab:ssl-baselines-vs-ours}
\end{table}

\subsection{Model Comparisons}
\paragraph{Zero-Shot GPT-4o vs. BERTweet-all (upper bound).} 
Table~\ref{tab:model_comparison} shows that GPT-4o achieves an averaged Macro-F1 of 0.628 on the training split and 0.641 on the test split across the 10 disaster events, slightly outperforming the GPT-4o results reported by \textcite{imran2024-openai}. These results provide an overall indication of the pseudo-label quality generated by GPT-4o. Analyzing performance by category over the combined training splits of all 10 events shows that GPT-4o performs well on clear and concrete categories, achieving Macro-F1 scores above 0.6. For instance, it reaches 0.885 on \textit{Injured or dead people} and 0.827 on \textit{Rescue, volunteering, or donation effort}. However, GPT-4o struggles on three categories (F1 $<$ 0.6), including \textit{Other relevant information} (0.276), \textit{Requests or urgent needs} (0.526), and \textit{Not humanitarian} (0.569), which are broader or less precise. For the remaining categories, the performance is 0.634 for \textit{Caution and advice}, 0.739 for \textit{Sympathy and support}, 0.766 for \textit{Displaced people and evacuations}, 0.698 for \textit{Missing or found people}, and 0.704 for \textit{Infrastructure and utility damage}. A detailed breakdown is provided in the project repository. Overall, categories with more accurate pseudo-labels are expected to benefit more from LLM-guided SSL approaches such as {\sc LG-CoTrain}.

Despite its relative good zero-shot performance, GPT-4o remains below the fully supervised BERTweet model trained on the complete training set, which achieves a Macro-F1 of 0.678 and serves as an approximate upper bound. This comparison suggests that while zero-shot LLMs offer competitive performance without task-specific training, further gains can be achieved through supervised or semi-supervised adaptation.

\textbf{VerifyMatch vs. classical SSL approaches. }
Table \ref{tab:ssl-baselines-vs-ours} shows that VerifyMatch yields competitive performance relative to the classical SSL baselines but lower ECE, especially at low 25--50 lb/cl budgets, indicating that combining LLM pseudo-labels with an explicit verification mechanism can improve robustness to noisy pseudo-labels. Compared to threshold-based or training-dynamics filtering strategies, VerifyMatch provides a more conservative but stable learning signal.

\textbf{\textsc{LG-CoTrain} vs. other SSL approaches.} Table~\ref{tab:ssl-baselines-vs-ours} also shows that under the most challenging conditions, {\sc LG-CoTrain} performs best among all SSL approaches: 0.608 Macro-F1 at 5 lb/cl and 0.619 at 10 lb/cl. This is a substantial improvement over classical SSL baselines such as ST, UST, and MixMatch, and also surpasses the best-performing baseline from \textcite{gupta_2025_calibrated} (Conf-ST-MixUp, AUM-ST-MixUp) and VerifyMatch in terms of Macro-F1 at both 5 and 10 lb/cl. These results suggest that incorporating LLM pseudo-labels within a co-training framework is particularly effective when the labeled set is too small for reliable self-training.

At higher label budgets, the performance gap between \textsc{LG-CoTrain} and other SSL baselines narrows, but with the average Macro F1 of 0.631 at 25 lb/cl, \textsc{LG-CoTrain} still outperforms the other SSL approaches. Notably, Self-Training becomes a strong baseline and achieves the best Macro-F1 value of 0.655 at 50 lb/cl. In this regime, the advantage of {\sc LG-CoTrain} diminishes, suggesting that once enough labeled data are available, the marginal benefit of LLM-guided pseudo-labeling is smaller. This may be due to the fact that the overall F1 obtained with GPT-4o is around 0.630-0.640. Furthermore, this can also be attributed to the limited amount of unlabeled data in the HumAID benchmark, and the fact that it may not be fully representative of minority classes, an issue that can also be encountered in real-world scenarios. Larger and more representative unlabeled datasets could help mitigate these limitations.

\textbf{GPT-4o vs. LLM-guided SSL approaches.} 
 In low-resource settings, {\sc LG-CoTrain} provides consistent gains over classical SSL methods on many events, which explains its strong averaged performance. 
While zero-shot GPT-4o remains highly competitive across events, {\sc LG-CoTrain} surpasses its performance in several cases. In particular, for the Hurricane Harvey, Irma, and Maria events under the 10, 25, and 50 labeled examples per class settings, {\sc LG-CoTrain} achieves higher Macro-F1 scores than GPT-4o. Even under the most constrained setting of 5 labeled examples per class, {\sc LG-CoTrain} still outperforms GPT-4o on Hurricane Harvey and Hurricane Maria.
This observation reinforces an important takeaway: when deployment constraints allow direct LLM inference, zero-shot LLMs can serve as strong baselines; however, when practitioners require a smaller deployable model, LLM-guided SSL can distill and transfer some of the LLM’s strengths into a compact classifier that can be executed effieciently and repeatedly at scale.

The amortized training paradigm of LLM-guided SSL approach such as {\sc LG Co-Train} is particularly advantageous in crisis-response settings, where streaming data volumes are high and rapid, low-latency and accurate decisions are required. Thus, the value of LLM-guided SSL lies not merely in marginal performance gains, but in enabling scalable, controllable, and cost-efficient deployment.

\textbf{SSL approaches vs. BERTweet (lower bound).} Also worth noting, all SSL approaches show better performance than the BERTweet trained only on the limited amount of labeled data, suggesting that SSL approaches represent a good option for classifying social media crisis data in a low-data regime.

\textbf{Calibration Behavior.} Beyond classification accuracy, calibration is critical in crisis-response applications, where confidence scores may influence downstream triage and operational decisions. Table~\ref{tab:ssl-baselines-vs-ours} shows that AUM-ST-MixUp achieves the lowest ECE at 10, 25 and 50 lb/cl, reflecting strong calibration under severe label scarcity. VerifyMatch also yields consistently low ECE values across all budgets, indicating that its verification mechanism can help control overconfident errors.
{\sc LG-CoTrain} prioritizes Macro-F1 improvements in the lowest-resource settings, while its calibration improves as more labeled data are available (ECE decreases from 0.174 at 5 lb/cl to 0.108 at 50 lb/cl).

Overall, the results highlight a practical trade-off: the most accurate method under extreme label scarcity is not always the best calibrated, and practitioners may need to balance performance and confidence reliability depending on operational needs.

\subsection{Per-event Analysis}

\begin{figure*}[t]
    \centering
    \includegraphics[width=\textwidth]{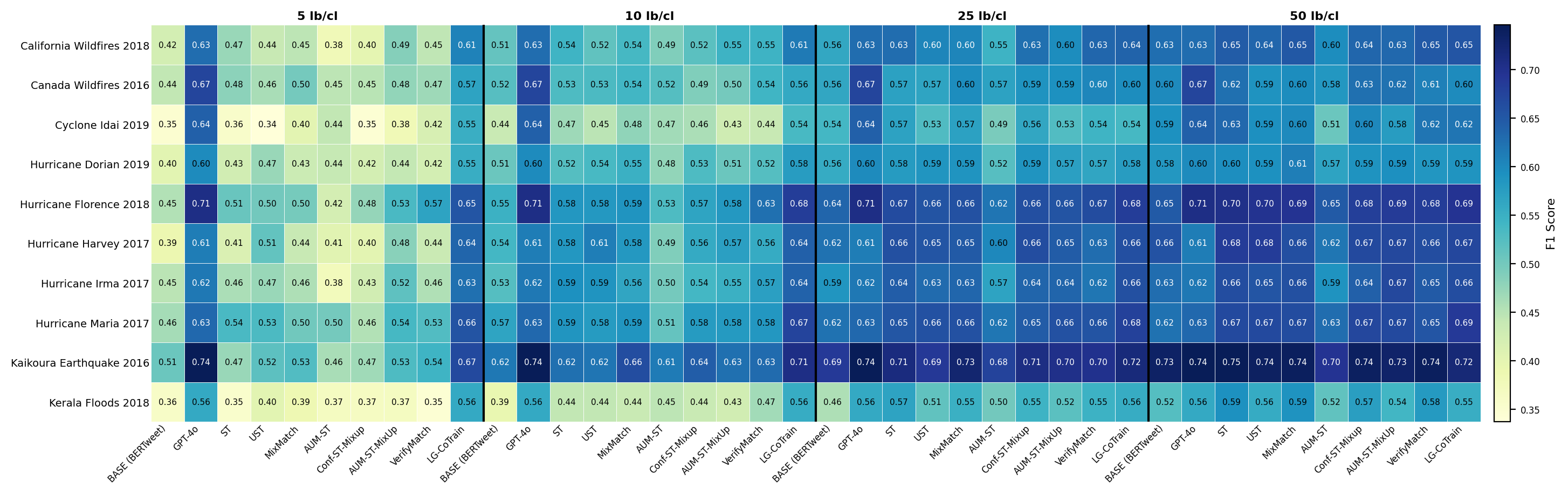}
    \caption{Per-event Macro-F1 scores for all methods across the 10 HumAID disaster events under different label budget settings. Values are rounded for readability.
    }
    \label{fig:macro_f1_heatmap}
\end{figure*}

Figure~\ref{fig:macro_f1_heatmap} presents per-event Macro-F1 scores for all methods across label budgets, providing additional insight beyond the averaged results. Performance varies substantially across disasters, reflecting differences in event characteristics and class distributions (shown in Table~\ref{tab:disaster_class_distribution}).

Similar to \textcite{imran2024-openai}, who find that LLMs struggle with flood events, we observe that both zero-shot GPT-4o and the SSL models perform relatively worse on the Kerala Floods 2018 event compared to others. We hypothesize that this variation may be attributable not only to the disaster type, but also largely to two factors: class imbalance and pseudo-label quality.
For events where a single class dominates the training set, such as Kerala Floods (53.8\% of training tweets belong to \textit{rescue\_volunteering\_or\_donation\_effort}) and Cyclone Idai (largest class 47.5\%), the models consistently underperform because rare classes have too few examples for reliable classification, and Macro-F1 penalizes poor recall on any single class equally. Conversely, for more balanced events such as Kaikoura Earthquake (largest class 22.5\%) and Hurricane Florence (largest class 23.6\%), the models achieve better scores overall as compared to the scores for other events. GPT-4o pseudo-label quality also plays a role: events where GPT-4o achieves higher zero-shot F1 tend to produce better {\sc LG-CoTrain} results, since higher-quality pseudo-labels provide a stronger training signal for co-training. The number of active classes also plays a role: events with only 8 classes (Canada Wildfires) or with extremely rare classes, such as Cyclone Idai where \textit{missing\_or\_found\_people} has only 13 training tweets, create near-zero F1 on those classes, which leads to a disproportionately reduced macro average.

\subsection{Ablation Study}

To quantify the contribution of the LLM in the LLM-guided SSL approaches, we compare {\sc LG-CoTrain} with GPT-4o pseudo-labels against the same co-training approach without LLM pseudo-labels, a variant called Self-guided CoTrain ~\textsc{(SG-CoTrain)}. ~\textsc{SG- CoTrain} replaces GPT-4o pseudo-labels with pseudo-labels from the BERTweet teacher model trained on the small labeled set. We run ~\textsc{(SG-CoTrain)} with 5 labeled examples per class across 10 events with 3 runs for each event. Table~\ref{tab:ablation} reports per-event Macro-F1 on the test set for these two experiments.

\begin{table}[h]
\centering
\small
\begin{tabular}{lccc}
\hline
\textbf{Event} & \textsc{LG-CoTrain} & \textsc{SG-CoTrain} & Delta \\
\hline
California Wildfires 2018      & 0.608 & 0.408 & $0.200$ \\
Canada Wildfires 2016          & 0.568 & 0.355 & $0.213$ \\
Cyclone Idai 2019              & 0.552 & 0.314 & $0.237$ \\
Hurricane Dorian 2019          & 0.554 & 0.409 & $0.145$ \\
Hurricane Florence 2018        & 0.655 & 0.319 & $0.336$ \\
Hurricane Harvey 2017          & 0.636 & 0.400 & $0.236$ \\
Hurricane Irma 2017            & 0.626 & 0.422 & $0.204$ \\
Hurricane Maria 2017           & 0.655 & 0.456 & $0.199$ \\
Kaikoura Earthquake 2016       & 0.669 & 0.501 & $0.168$ \\
Kerala Floods 2018             & 0.560 & 0.378 & $0.182$ \\
\hline
\textbf{Average}               & \textbf{0.608} & \textbf{0.396} & $\mathbf{0.212}$ \\
\hline
\end{tabular}
\caption{Ablation study for co-training with 5 labels for class, with LLM pseudo-labels ({\sc LG-CoTrain}) and without LLM pseudo-labels ({\sc SG-CoTrain}). The Delta improvement obtained from the LLM pseudo-labels is also shown.}
\label{tab:ablation}
\end{table}

{\sc LG-CoTrain} consistently outperforms {\sc SG-CoTrain} across all events, with gains ranging from 0.145 to 0.336 in F1 (average improvement of 0.212). Since both models share the same co-training pipeline and hyperparameter search space, this performance gap can be attributed primarily to differences in pseudo-label quality. This result is expected: a BERTweet teacher trained on only 50 labeled examples (5 lb/cl) produces pseudo-labels that are too noisy for effective co-training and even degrade performance through error propagation. In contrast, GPT-4o’s zero-shot predictions are sufficiently accurate to provide a strong initial signal, enabling more effective learning.

To further examine {\sc SG-CoTrain}, we also run an simple experiment using fixed hyper-parameter for all events with filtering pseudo-labeled data from the BERTweet teacher by retaining only the 50 most confident predictions per class. That preliminary results show that applying confidence filtering improves performance, despite using fewer pseudo-labeled samples. And we also observe the gap between {\sc LG-CoTrain} and non-LLM variants tends to narrow as the amount of labeled data increases, which suggests that with sufficient labeled data, confidence-filtered self-guidance can partially substitute for LLM guidance. But we will leave the verification of this hypothesis and more comprehensive analysis to future work.

\subsection{Deployment Considerations}

In our experiments, GPT-4o zero-shot classification of the evaluation set (6,463 tweets across 10 events) cost \$17.83 (via the OpenAI Batch API). The unlabeled training pool totals 44,373 tweets (approximately 4,400 per event on average), so we estimate pseudo-labeling cost at roughly \$12 per event at this scale. Including hyperparameter tuning and model training on a dataset of this size, the full pipeline for a single new disaster event requires under one hour on an 8-GPU NVIDIA H100 cluster, cost of which could start from \$24 depending on the chosen cloud platform. The total deployment cost is therefore approximately \$36 per event. More affordable GPU may reduce this cost but would require more training time. These estimates are based on our dataset with an average tweet length typical of Twitter/X posts; actual costs will vary with the number of tweets, their average length, as well as the LLM API cost per token. 

Taken together, our experimental results suggest three practical implications. First, LLM-guided SSL (especially {\sc LG-CoTrain}) is most beneficial when labeled data are extremely scarce (5--10 lb/cl), although the cost of using LLMs should be carefully considered in real-world settings with large volumes of noisy social media data. Second, as the amount of labeled data increases, simpler SSL methods such as Self-Training become highly competitive and can even outperform more complex approaches, making them strong baselines in moderate-resource settings. Third, calibration varies substantially across methods, and approaches such as AUM-ST-MixUp and VerifyMatch may be preferred when reliable confidence estimates are important.

In practice, this suggests the following. When only a very small labeled set is available (e.g., 5 lb/cl) and rapid deployment is needed, a cost-effective strategy is to generate a limited amount of pseudo-labeled data with an LLM and then train an {\sc LG-CoTrain} model to filter actionable social media content. When a moderate amount of labeled data is available (e.g., 50 lb/cl), it may be preferable to avoid LLM costs and instead rely on task-specific SSL methods such as Self-Training. Prior work such as \textcite{guo_2025_asonam} also shows that task-specific models are often preferred when feasible, as they can outperform LLMs in zero-shot settings.
 Finally, when interpretability and well-calibrated confidence estimates are critical, methods such as AUM-ST-MixUp provide a more suitable choice.

\section{Conclusion and Future Work}

This paper provides the first empirical evaluation of LLM-guided semi-supervised learning for social media crisis classification on the HumAID benchmark. We compare two recent LLM-assisted SSL methods, VerifyMatch and {\sc LG-CoTrain}, against widely used SSL baselines under multiple label budgets and evaluate the results in terms of predictive performance (Macro-F1) and as well as reliability (ECE).

Our results show that {\sc LG-CoTrain} achieves the strongest Macro-F1 in extremely low-resource settings (5--10 labeled examples per class), demonstrating that LLM pseudo-labels can meaningfully improve SSL when labeled data are severely limited. VerifyMatch provides competitive classification performance and strong calibration, while AUM-ST-MixUp remains a strong choice when calibration is the primary concern. As the label budget increases, the performance advantage of LLM-guided SSL decreases and Self-Training becomes a difficult-to-beat baseline, emphasizing that method choice should reflect the available annotation budget and deployment requirements. We discuss corresponding recommendations for practical deployment.

Future work will explore three directions. First, we will investigate how unlabeled data scale and representativeness affect LLM-guided SSL, especially under missing-class or distribution-shift scenarios that can arise in real operations. Second, we will study calibration-aware training objectives and pseudo-label filtering strategies tailored to crisis informatics, aiming to improve both accuracy and reliability. Finally, we will extend the evaluation to multimodal crisis datasets and cross-event generalization settings, where LLMs may provide even greater benefits as a source of transferable knowledge. 

\section{Acknowledgment}
This work is supported by a collaborative CAHSI-Google Institutional Research Program award.

\printbibliography

\end{document}